# Deep Reinforcement Learning for Weapons to Targets Assignment in a Hypersonic strike

Brian Gaudet*
*University of Arizona, 1127 E. Roger Way, Tucson Arizona, 85721*

Kristofer Drozd[†]
*University of Arizona, 1127 E. Roger Way, Tucson Arizona, 85721*

Roberto Furfaro[‡]
*University of Arizona, 1127 E. Roger Way, Tucson Arizona, 85721*

We use deep reinforcement learning (RL) to optimize a weapons to target assignment (WTA) policy for multi-vehicle hypersonic strike against multiple targets. The objective is to maximize the total value of destroyed targets in each episode. Each randomly generated episode varies the number and initial conditions of the hypersonic strike weapons (HSW) and targets, the value distribution of the targets, and the probability of a HSW being intercepted. We compare the performance of this WTA policy to that of a benchmark WTA policy derived using non-linear integer programming (NLIP), and find that the RL WTA policy gives near optimal performance with a 1000X speedup in computation time, allowing real time operation that facilitates autonomous decision making in the mission end game.

## I. Introduction

The ability to implement effective autonomous hypersonic strikes against an adversary can potentially change the balance of power for our forces located in proximity to a powerful adversary, significantly extending the depth of our defense. Hypersonic weapons have an advantage over slower missiles in that they both reduce the response time of a strike and are more difficult to intercept. The fastest response time for weapons deployed at large distances from the theater of operations would be achieved by boost-glide hypersonic strike weapons (HSW) or maneuverable ballistic re-entry vehicles (MBRV) launched on a depressed trajectory. Both approaches could be made more economical by launching multiple HSW or MBRV from a single rocket, similar to our approach with our nuclear deterrent. Hypersonic cruise missiles launched from strategic bombers would have slower response times, but may be more economical. In the following, HSW will refer to either a maneuverable hypersonic vehicle capable of following non-ballistic trajectories and MBRV launched on depressed ballistic trajectories but capable of maneuvering in the terminal mission phase. In both cases, we assume the HSW has target discrimination and imaging capability, allowing relative navigation and guidance to a selected target.

There are multiple problems that need to be solved to make autonomous hypersonic strikes possible. First, in order to achieve small miss distances (< 5m) and estimate the value of potential targets, the HSW needs to image targets while operating in the hypersonic flight regime. The radome material must be both transparent to radar and able to withstand the high temperature and pressure of low altitude hypersonic flight, and the extreme heat may ionize the boundary layer and create interference. Assuming the target can be imaged, another problem (that we take a step towards solving in this paper) is that there might be limited real-time information on the exact location and composition of mobile enemy forces, requiring an effective hypersonic strike to be carried out with full autonomy (as opposed to a pre-computed targeting plan). Other problems include inter-vehicle communication, real-time guidance that can satisfy heating rate and load constraints (briefly discussed in II.B), and robust and adaptive flight control [1, 2].

Real-time, autonomous cooperation between HSW is required to maximize strike effectiveness, which we quantify as the total value of destroyed targets. One approach to maximizing strike effectiveness is to formulate the multi-agent problem as a weapons to target assignment (WTA) problem that takes into account both the value of potential targets and the probability that a given HSW can reach a target without being intercepted. Given that the other problems previously posed can be solved, WTA will have a significant impact on mission effectiveness for the portion of HSW that survive to the terminal mission phase; that is, the phase where the vehicles can image targets with their sensors. The WTA problem is essentially a non linear integer programming (NLIP) problem [3], assigning some number of HSW to each target, with the objective of maximizing the total value of destroyed targets. Although we can formulate the WTA problem in the NLIP framework, we later show that the computation time (even on a high performance computer) is too slow for real-time operation in the hypersonic strike scenario, where the range limitation of on-board sensors result in a terminal phase (where the HSW can image the target, allowing navigation and guidance) lasting from 8 to 16 seconds. Moreover, the NLIP solver runs out of memory for large numbers of HSW and targets. Thus, there is a need for a WTA policy that can be computed in real time and can scale to large numbers of HSW and targets.

*Research Engineer, Department of Systems and Industrial Engineering, E-mail: briangaudet@arizona.edu
[†]PhD Student, Department of Systems and Industrial Engineering, E-mail: kdrozd@arizona.edu
[‡]Professor, Department of Systems and Industrial Engineering, Department of Aerospace and Mechanical Engineering. E-mail: robertof@arizona.edu

Previous work on the WTA problem include [4], an overview of different approaches to the problem. In [5], the authors address a scenario with multiple targets, defenders, and missiles (which is similar to the scenario addressed in this work) and use an adjoint system representation to evaluate the merit of a particular assignment of weapons to targets. However, the computation takes around an hour, and is therefore unsuited for real time implementation for a hypersonic strike. A WTA solver that uses a modified cost function to insure simultaneous arrival of weapons to targets is described in [6], but does not take into account target defensive measures, is not compared to an optimal solution, and is probably too computationally expensive for the hypersonic strike application. in [7], the authors compare different cost functions and optimization algorithms, taking into account weapon attrition and target feasibility. Finally, in [8], the authors use reinforcement learning to solve a static WTA problem in a defensive application. Their WTA policy is shown to have performance exceeding a non-optimal benchmark, but although the policy can generalize to varying numbers of targets, it cannot do so for varying numbers of weapons.

In this manuscript, we show that deep reinforcement learning (RL) can be used to learn an effective WTA policy that can be computed in real time (a few milliseconds). This opens the possibility of *dynamic* WTA, where the WTA is updated periodically during the terminal phase. For example, if an HSW assigned to a high value target is intercepted early in the engagement, an HSW assigned to a lower value target may have a feasible trajectory to the high value target, and could be diverted to replace the destroyed HSW. Another advantage of our RL approach to WTA is that the computation time scales approximately linearly with the number of HSW and targets ($n \times m$). In prior work[2, 9–16] we have shown that both RL and meta-reinforcement learning can be successfully applied to aerospace guidance, navigation, and control problems. To our knowledge, this work is the first application of RL to the WTA problem that uses a convolutional network (CNN) [17] front end, compares performance to an optimal benchmark, addresses the hypersonic strike application, and is compatible with a variable number of HSW and targets. The remainder of the paper is organized as follows. Section II formulates the WTA problem, Section III describes the methods used to develop both the NLIP benchmark WTA and the RL WTA, Section IV compares the performance of the NLIP WTA, RL WTA, and a simple heuristic WTA with a fast run time. Conclusions and future work are given in Section VI.

## II. Problem Formulation

### A. Episode Initialization

The simulator models a hypersonic strike scenario where $m$ HSW engage in a hypersonic strike against $n$ targets, and $m$ and $n$ are randomized at the start of each episode within bounds dictated by the capability of the benchmark NLIP solver. The HSW positions $\mathbf{r}_M$ are initialized by randomly selecting a distance, elevation angle, and azimuth angle, all referenced to points (one point per HSW) inside a radius $\mathcal{R}_\text{expected}$. The circle defined by this radius represents an area where the targets are expected to be found. The targeted points within the circle could represent either a random grid (no knowledge of target distribution) or the most likely target locations based on intelligence, surveillance, and reconnaissance (ISR). The mission is designed so that when the WTA is computed, all HSW are at roughly the same distance to their initial targets, with the distance equal to the sensor range limit associated with reliable target discrimination[*]. We assume an effective sensor range of 30 km due to the limited power likely to be available to active radar on a HSW. At the start of each episode, we re-initialize the HSW positions until the minimum HSW spacing falls above the bound shown in Table 1 Lower initial altitudes are associated with smaller initial speeds.

Targets are uniformly spaced within a circle of radius $\mathcal{R}_\text{GT}$ and have a randomized initial velocity $\mathbf{v}_M$ (obviously planar). $\mathcal{R}_\text{GT}$ is independently sampled from $\mathcal{R}_\text{expected}$. Thus, there can be considerable distance between where the targets are and where they are expected to be. Each target is then shifted in a random direction and distance $d_\text{shift}$, which represents errors in ISR and target motion between the ISR mission and the strike mission. Targets engage in jinking maneuvers with acceleration $\mathbf{a}_T$. The target values $\mathcal{V}$ are randomly drawn from a distribution with number of values $\mathcal{V}_n$ and value bounds $\mathcal{V}_{\min}, \mathcal{V}_{\max}$, randomly generated as shown in Table 2. Values are sampled with probability weights inversely proportional to the values.

---
[*]The HSW would likely be integrating and sharing sensor measurements for a short time prior to reaching this distance

Table 1  HSW Initial Conditions

| Parameters Drawn Uniformly | Min | Max |
|---|---|---|
| Radius of circle containing initial HSW targets (m) $\mathcal{R}_{\text{expected}}$ | 6000 | 12000 |
| Distance $d$ (m) | 30000 | 30000 |
| Azimuth $\phi$ (degrees) | -45 | 45 |
| Elevation $\theta$ (degrees) | 45 | 90 |
| Velocity Magnitude $\|\mathbf{v}_M\|$ (m/s) | 2500 | 3500 |
| Minimum HSW Spacing (m) | 2000 | 2000 |

Table 2  Target Initial Conditions

| Parameters Drawn Uniformly | Min | Max |
|---|---|---|
| Radius $\mathcal{R}_{\text{GT}}$ (m) | 6000 | 12000 |
| Target shift $d_{\text{shift}}$ (m) | 1000 | 3000 |
| Velocity Magnitude $\|\mathbf{v}_T\|$ (m/s) | 0 | 30 |
| Acceleration Magnitude $\|\mathbf{a}_T\|$ (m/s$^2$) | 0 | 1 |
| Number of values $\mathcal{V}_n$ | 3 | 5 |
| Minimum Value Bound $\mathcal{V}_{\min}$ | 1 | 1 |
| Maximum Value Bounds $\mathcal{V}_{\max}$ | 10 | 15 |

## B. Guidance

In an actual application, the guidance system would need to implement path constraints on heating rate, load, and dynamic pressure, which would be used to determine the feasible set of targets. Here, we simplify the problem and use proportional navigation (PN) for guidance, with heading error used as a proxy for feasibility, assuming that a heading error HE of less than 15 degrees results in a feasible trajectory. We verified (by setting the threat model probability of interception to zero) that HE < 20° results in miss distances of less than 5m, but use HE < 15° to account for the impact of path constraints. Note that it may be possible to use PN in a deployed implementation. PN can be used to satisfy load constraints, and at least for the case of a MBRV, an unguided implementation can obviously satisfy the required heating rate and dynamic pressure constraints. Adding guidance to an MBRV will induce lateral acceleration, which will reduce speed, and therefore heating rate and dynamic pressure. However, it would also be possible to explicitly consider path constraints through a feasibility metric that maps time to go, heading error, altitude, speed, and acceleration capability to a binary feasibility variable. Although it is unlikely there is a closed form solution for such a metric that takes into account drag, it may be possible to learn this function through simulated experience using a high fidelity 6-DOF aerodynamic model. We leave this to future work.

Our implementation of PN guidance is shown in Eqs. 1a through 1c, where $\mathbf{r}_{\text{TM}} = \mathbf{r}_T - \mathbf{r}_M$, $\mathbf{v}_{\text{TM}} = \mathbf{v}_T - \mathbf{v}_M$, $\mathbf{a}_M$ is the achieved HSW acceleration, and $v_c = -\dfrac{\mathbf{r}_{\text{TM}} \cdot \mathbf{v}_{\text{TM}}}{\|\mathbf{r}_{\text{TM}}\|}$. We use $N = 5$ and $a_{\max} = 40$ g, where g = 9.81m/s$^2$.

$$\mathbf{\Omega} = \frac{\mathbf{r}_{\text{TM}} \times \mathbf{v}_{\text{TM}}}{\mathbf{r}_{\text{TM}} \cdot \mathbf{r}_{\text{TM}}} \tag{1a}$$

$$\mathbf{a}_{\text{com}} = -N v_c \left( \frac{\mathbf{r}_{\text{TM}}}{\|\mathbf{r}_{\text{TM}}\|} \times \mathbf{\Omega} \right) \tag{1b}$$

$$a = \text{clip}\left(\|\mathbf{a}_{\text{com}}\|, 0, a_{\max}\right) \tag{1c}$$

$$\mathbf{a}_{\text{M}} = a \frac{\mathbf{a}_{\text{com}}}{\|\mathbf{a}_{\text{com}}\|} \tag{1d}$$

### C. Sensor Model

We use a simplified sensor model that, regardless of range and look angle to target, has a 10% chance of replacing the true target value with a target value either one class higher or one class lower. For example, if in a given episode the possible target values were [1, 3, 5, 9] and the ground truth value was 5, the observed value given to the WTA policy would be 3 with probability 0.05 and 9 with probability 0.05. A deployed system could implement active electronically scanned array (AESA) radar, which would allow simultaneous search, tracking, and discrimination of targets covered by a large field of regard. Although synthetic aperture radar would be more accurate, it might be too slow for the hypersonic strike application. Future work will implement a more accurate sensor model that takes into account both range to target and look angle.

### D. Threat Model

We use a simplified stochastic threat model that targets a given HSW with a stochastic targeting rate of $p_{\text{targeted\_per\_sec}}$, and once targeted, the HSW is intercepted with a stochastic interception rate of $p_{\text{intercepted\_per\_sec}}$. These rates are converted to per-step rates by multiplying by the navigation period of 0.1 s. At each step, the HSW is deemed targeted if $\mathcal{U}(0, 1, 1) < p_{\text{targeted\_per\_step}}$ and once targeted, is deemed intercepted if $\mathcal{U}(0, 1, 1) < p_{\text{intercepted\_per\_step}}$. Here $\mathcal{U}(a, b, n)$ denotes an $n$ dimensional uniformly distributed random variable bounded by $(a, b)$ that is generated at each step, and each dimension of the random variable is independent. In reality, due to the greater range of ship or land deployed radar compared to airborne radar, there would be significant attrition before the strike's terminal phase.

**Table 3   Threat Model**

| Parameters Drawn Uniformly | Min | Max |
| --- | --- | --- |
| Targeting Rate $p_{\text{targeted\_per\_sec}}$ (m) | 0.2 | 0.3 |
| Interception Rate $p_{\text{intercept\_per\_sec}}$ (s$^{-1}$) | 0.1 | 0.3 |

### E. Equations of Motion

The HSW inertial frame position $\mathbf{r}_{\text{M}}$ and velocity unadjusted for drag $\tilde{\mathbf{v}}_{\text{M}}$ are updated by integrating Eqs. (2a) and (2b). The HSW speed $V_{\text{M}} = \|\mathbf{v}_{\text{M}}\|$ is adjusted for drag by integrating Eq. (2c), where $k_{\text{M}} = 0.1$, $\text{cd}_0 = 0.1$, and $\rho$ is the atmospheric pressure at altitude $h_{\text{M}}$. We use $m_{\text{M}} = 450$ kg.

$$\dot{\mathbf{r}}_{\text{M}} = \mathbf{v}_{\text{M}} \tag{2a}$$

$$\dot{\tilde{\mathbf{v}}}_{\text{M}} = \mathbf{a}_{\text{M}} \tag{2b}$$

$$\dot{V}_{\text{M}} = -\frac{\rho(h_{\text{M}}) V_{\text{M}}^2 \text{cd}_0}{2 m_{\text{M}}} - k_{\text{M}} \|[\mathbf{a}_{\text{M}}]\| \tag{2c}$$

The threat velocity is then computed as shown in Eq. (3)

$$\mathbf{v}_{\text{M}} = V_{\text{M}} \frac{\tilde{\mathbf{v}}_{\text{M}}}{\|\tilde{\mathbf{v}}_{\text{M}}\|} \tag{3}$$

The equations of motion are updated using fourth order Runge-Kutta integration. For ranges greater than 300m, a timestep of 0.1s is used, and for the final 300m of homing, a timestep of 0.01s is used in order to more accurately (within 1m) measure miss distance; this technique is borrowed from [18].

# III. Methods

## A. NLIP WTA

The ideal assignment of multiple hypersonic strike weapons (HSW) can be represented as a NLIP problem as shown in Eqs. (4) and (5):

$$\text{Maximize} \sum_{j=1}^{N_t} c_j \left(1 - \prod_{i=1}^{N_v} \frac{F_{ij}}{F_{ij}^{(1-x_{ij})}}\right), \tag{4}$$

subject to

$$\sum_{j=1}^{N_t} x_{ij} = 1 \quad (i = 1, 2, \ldots, N_t). \tag{5}$$

The variable $x_{ij}$ is binary and defined as shown in Eq. (6).

$$x_{ij} = \begin{cases} 1 & \text{if vehicle } i \text{ is assigned to target } j, \\ 0 & \text{otherwise.} \end{cases} \tag{6}$$

The equality constraint specifies that a vehicle can only be assigned to one target, but a target can have many vehicles assigned to it. The variable $c_j$ is the value of target $j$ being destroyed. The variable $F_{ij}$ is a cumulative distribution function (CDF) that gives the probability of vehicle $i$ being intercepted if it is assigned to target $j$. The product inside the objective function gives the probability of all vehicles assigned to target $j$ being intercepted. Subtracting the product from 1 gives the probability that at least one vehicle is not intercepted and the target is destroyed.

The CDF depends on the time of flight $t_{f_{ij}}$ required to guide vehicle $i$ to target $j$. The time of flight can be estimated ahead of time via Eq. (7), where $\mathbf{r}_{\text{TM}_0}$ is the vector from the initial position of the vehicle to the target and $v_{c_0}$ is the magnitude of the vehicles initial velocity projected on $\mathbf{r}_{\text{TM}_0}$,

$$t_{f_{ij}} = \frac{\|\mathbf{r}_{\text{TM}_0}\|_2}{v_{c_0}}. \tag{7}$$

The probability of interception is modeled by the summation of two exponential distributions given in Eq. (8), where $T_1$ is a random variable for the time until a vehicle is targeted, $T_2$ is a random variable for the time until a vehicle is intercepted once targeted, $\lambda_1$ is the target rate, and $\lambda_2$ is the interception rate.

$$F_{ij}(t_{f_{ij}}) = P(T_1 + T_2 \leq t_{f_{ij}}) = 1 - e^{-\lambda_1 t_{f_{ij}}} - \lambda_2 t_{f_{ij}} e^{-\lambda_2 t_{f_{ij}}}, \tag{8}$$

Both the targeting and interception rates are stochastic, so we use the expected value of these rates, which in our case are 0.25 s$^{-1}$ and 0.2 s$^{-1}$, respectively (see Section II.D). Thus, the mean of $T_1$ and $T_2$ are 4 s and 5 s, respectively. We incorporate feasibility by setting $F_{ij} = 0.9999$ if the initial heading error between HSW $i$ and target $j$ is greater than 15°. Setting $F_{ij}$ very close to 1 means there is a negligible chance vehicle $i$ reaches target $j$, which is the case when the initial heading error is above 15°. The NLIP problem just described is solved using the open source software package GEKKO [19] for Python. For instances where GEKKO failed to find a solution, which was rare, vehicles were assigned to the target that resulted in the lowest initial heading error.

## B. Reinforcement Learning for WTA

### 1. The Reinforcement Learning Framework

In the reinforcement learning framework, an agent learns through episodic interaction with an environment how to successfully complete a task using a policy that maps observations $\mathbf{o}$ to actions $\mathbf{u}$. The environment initializes an episode by randomly generating a ground truth state $\mathbf{x}$, mapping this state to an observation, and passing the observation to the agent. The agent uses this observation to generate an action that is sent to the environment. The environment then uses the action and the current ground truth state to generate the next state and a scalar reward signal $r(\mathbf{x}, \mathbf{u})$. The reward and the observation corresponding to the next state are then passed to the agent. The process repeats until the environment terminates the episode, with the termination signaled to the agent via a done signal. Trajectories collected over a set of episodes (referred to as rollouts) are collected during interaction between the agent and environment, and used to update the policy and value functions.

The PPO algorithm [20] used in this work is a policy gradient algorithm which has demonstrated state-of-the-art performance for many reinforcement learning benchmark problems. PPO approximates the Trust Region Policy Optimization method [21] by accounting for the policy adjustment constraint with a clipped objective function. The objective function used with PPO can be expressed in terms of the probability ratio $p_k(\theta)$ given by,

$$p_k(\theta) = \frac{\pi_\theta(\mathbf{u}_k|\mathbf{o}_k)}{\pi_{\theta_{\text{old}}}(\mathbf{u}_k|\mathbf{o}_k)} \tag{9}$$

The PPO objective function is shown in Equations (10a) through (10c). The general idea is to create two surrogate objectives, the first being the probability ratio $p_k(\theta)$ multiplied by the advantages $A_\mathbf{w}^\pi(\mathbf{o}_k, \mathbf{u}_k)$ (see Eq. (11)), and the second a clipped (using clipping parameter $\epsilon$) version of $p_k(\theta)$ multiplied by $A_\mathbf{w}^\pi(\mathbf{o}_k, \mathbf{u}_k)$. The objective to be maximized $J(\theta)$ is then the expectation under the trajectories induced by the policy of the lesser of these two surrogate objectives.

$$\text{obj1} = p_k(\theta) A_\mathbf{w}^\pi(\mathbf{o}_k, \mathbf{u}_k) \tag{10a}$$
$$\text{obj2} = \text{clip}(p_k(\theta) A_\mathbf{w}^\pi(\mathbf{o}_k, \mathbf{u}_k), 1-\epsilon, 1+\epsilon) \tag{10b}$$
$$J(\theta) = \mathbb{E}_{p(\tau)}[\min(\text{obj1}, \text{obj2})] \tag{10c}$$

This clipped objective function has been shown to maintain a bounded Kullback-Leibler (KL) divergence [22] with respect to the policy distributions between updates, which aids convergence by ensuring that the policy does not change drastically between updates. Our implementation of PPO uses an approximation to the advantage function that is the difference between the empirical return and a state value function baseline, as shown in Equation 11, where $\gamma$ is a discount rate and $r$ the reward function.

$$A_\mathbf{w}^\pi(\mathbf{x}_k, \mathbf{u}_k) = \left[\sum_{\ell=k}^{T} \gamma^{\ell-k} r(\mathbf{x}_\ell, \mathbf{u}_\ell)\right] - V_\mathbf{w}^\pi(\mathbf{x}_k) \tag{11}$$

Here the value function $V_\mathbf{w}^\pi$ is learned using the cost function given by

$$L(\mathbf{w}) = \frac{1}{2M} \sum_{i=1}^{M} \left(V_\mathbf{w}^\pi(\mathbf{x}_k^i) - \left[\sum_{\ell=k}^{T} \gamma^{\ell-k} r(\mathbf{u}_\ell^i, \mathbf{x}_\ell^i)\right]\right)^2 \tag{12}$$

In practice, policy gradient algorithms update the policy using a batch of trajectories (roll-outs) collected by interaction with the environment. Each trajectory is associated with a single episode, with a sample from a trajectory collected at step $k$ consisting of observation $\mathbf{o}_k$, action $\mathbf{u}_k$, and reward $r_k(\mathbf{o}_k, \mathbf{u}_k)$. Finally, gradient ascent is performed on $\theta$ and gradient descent on $\mathbf{w}$, with their update equations given by

$$\mathbf{w}^+ = \mathbf{w}^- - \beta_\mathbf{w} \nabla_\mathbf{w} L(\mathbf{w})|_{\mathbf{w}=\mathbf{w}^-} \tag{13}$$
$$\theta^+ = \theta^- + \beta_\theta \nabla_\theta J(\theta)|_{\theta=\theta^-} \tag{14}$$

where $\beta_\mathbf{w}$ and $\beta_\theta$ are the learning rates for the value function, $V_\mathbf{w}^\pi(\mathbf{o}_k)$, and policy, $\pi_\theta(\mathbf{u}_k|\mathbf{o}_k)$, respectively.

### 2. RL applied to the WTA problem

We represent engagements allowing a maximum number of $m$ HSW and $n$ targets using a matrix $\mathbf{E} \in \mathbb{R}^{m \times n \times k}$, where the first of the $k$ channels are allocated to the feasibility[†] $\mathcal{F}$ of assigning HSW $i$ to target $j$ (1 if feasible, -1 if infeasible), the second to the time-to-go $t_{\text{go}} = \frac{\|\mathbf{r}_{\text{TM}}\|}{v_c}$ from HSW $i$ to target $j$, and the third channel to the observed value $\mathcal{V}$ of target $j$ from target $i$[‡]. An assignment can be invalid if it involves an HSW or target index higher than the actual number in an engagement. For example, if the maximum number of HSW and targets is 20 and 12, then any assignment with $i > 12$ or $j > 10$ would be invalid, and the channel entries for $\mathcal{F}$, $t_{go}$, and $\mathcal{V}$ would be replaced by -1. Similarly, if HSW $i$ has been intercepted or target $j$ destroyed, or HSW $i$ assigned target $j$ is infeasible, the channel entries are replaced by -1.

---
[†]Recall that in this work, we use a heading error < 15° as a proxy for feasibility
[‡]Although our current sensor model (Section II.C) is not range or look angle dependent and is thus independent of the observing HSW, for a higher fidelity sensor model this would be the case

In our experiments, we found that using a convolutional network [17] to process the engagement state $\mathbf{E}$ gave faster run time and higher performance than flattening $\mathbf{E}$ and using a fully connected network architecture. The observation $\mathbf{E}$ is passed to two parallel convolutional networks, one for the policy network and another for the value function network. Each convolutional network has two stages. The first stage has 3 input channels, 8 filters of size=3, padding=1 and stride=1, and the second stage has 8 input channels, 8 filters of size=3, padding=1, and stride=2. Note we do not use any pooling layers, but instead take the approach described in [23]. Each stage uses the rectified linear (RELU) activation function [24]. The output of the second stage is flattened and passed to three fully connected layers which differ between the policy and value function networks, and are shown in Table 4, where $z$ denotes the dimension of the flattened output of the convolutional layer, $n_{hi}$ is the number of units in layer $i$, and act_dim is the action dimension.

Table 4   Policy and Value Function fully connected layers

| Layer | Policy Network # units | activation | Value Network # units | activation |
|---|---|---|---|---|
| hidden 1 | $\sqrt{z * n_{h3}}$ | tanh | $\sqrt{z * n_{h3}}$ | tanh |
| hidden 2 | $10 * \text{act\_dim}$ | tanh | 5 | tanh |
| output | act_dim | linear | 1 | linear |

In our PPO implementation, we set the clipping parameter $\epsilon$ to 0.1. The policy and value function are learned concurrently, as the estimated value of a state is policy dependent. The policy uses a multi-categorical policy distribution, where a separate observation conditional categorical distribution is maintained for each element of the action vector $\mathbf{a} \in \mathbb{Z}^m$, where in this application, we have $n$ possible actions (target assignments) for each element of the action vector. Specifically, the action distribution (the WTA) is implemented by applying the softmax function to the $n$ logits (a logit is an output of the policy network) corresponding to each element of the action vector, as shown in Equation (15), where $p(a_{ij})$ is the probability of taking action $j \in \{0, m-1\}$ for the $i^{th}$ element of the action vector (here $i \in \{0, n-1\}$) and $z_{ij}$ is the logit corresponding to action $j$.

$$p(a_{ij}|\mathbf{E}) = \frac{e^{z_{ij}|\mathbf{o}}}{\sum_j e^{z_{ij}|\mathbf{E}}} \tag{15}$$

During optimization, the policy samples from this distribution, returning a value $\mathbf{u} \in \mathbb{Z}$, $u_i \in \{0, m-1\}, i \in \{0, n-1\}$. For testing and deployment, the sampling is turned off, and the action is just the argmax of the $m$ logits across each element of the action vector. Note that exploration is conditioned on the observation, with the $m$ logits associated with each element of the action vector determining how peaked the softmax distribution becomes for each action. Because the probabilities in Equation 9 are calculated using the logits, the degree of exploration automatically adapts during learning such that the objective function is maximized.

The reward function is shown in Eq. (16), where $r$ is the reward, $\alpha$ the reward coefficient, $M$ is the number of targets in the episode, $\mathcal{V}_i$ the ground truth value of target $i$, and $d_i$ is a Boolean variable set True of the target $i$ is destroyed during the episode. The reward is given at the end of an episode, and we use $\alpha = 10$. It is worth pointing out that from the perspective of the RL optimization framework, each episode has only one step. The agent observes the engagement state at the start of the episode and takes an action. The environment then runs the episode (which has many steps, see Section II.E) and at the end returns the reward to the agent. In other words, this is actually a contextual bandit problem rather than a Markov Decision Process.

$$r = \alpha \sum_i^M d_i \mathcal{V}_i \tag{16}$$

## IV. Experiments

The following results are for a static WTA, computed at the start of each episode. We experimented with different types of dynamic WTA, including updating the WTA at fixed times or when a HSW is destroyed, but found that the results from both the RL and benchmark methods were slightly worse than when a static WTA was performed. We attribute this to our use of heading error as a proxy for feasibility, and the fact that for a given HSW, the heading error to multiple targets can initially increase early in the engagement (due to decreasing range) before the guidance acceleration reduces the heading error to the current target. This issue could be solved by a better proxy for feasibility as discussed in Section II.B. Alternatively, if an optimal control guidance law could be designed to run fast enough for the dynamic WTA problem (unlikely), it could be used to determine feasibility. In Section V we discuss how the current RL WTA policy can be used in a dynamic WTA application using the same feasibility criterion as used in these experiments . The NLIP solver ran out of memory when the number of HSW and targets were increased beyond 20 and 12, respectively. Therefore, to allow comparability, we set the maximum number of HSW and targets in an episode to 20 and 12.

### A. RL WTA Optimization Results

We found that the best performance at test time was achieved using an RL WTA that was optimized with the number of HSW and targets fixed at their maximum values, 20 and 12 respectively. However, we tested with randomized numbers of HSW and targets. Learning curves are illustrated in Fig. 1, where we see that occasionally the reward is zero due to all the HSW being intercepted. The statistics are computed over the rollouts (2000 episodes per rollout). We used $\epsilon = 0.1$ in Eq. 10b, and the Adam optimizer [25] for the adaptive learning rate. Since PPO is stochastic, we ran four parallel optimization runs and chose the policy with the best test performance.

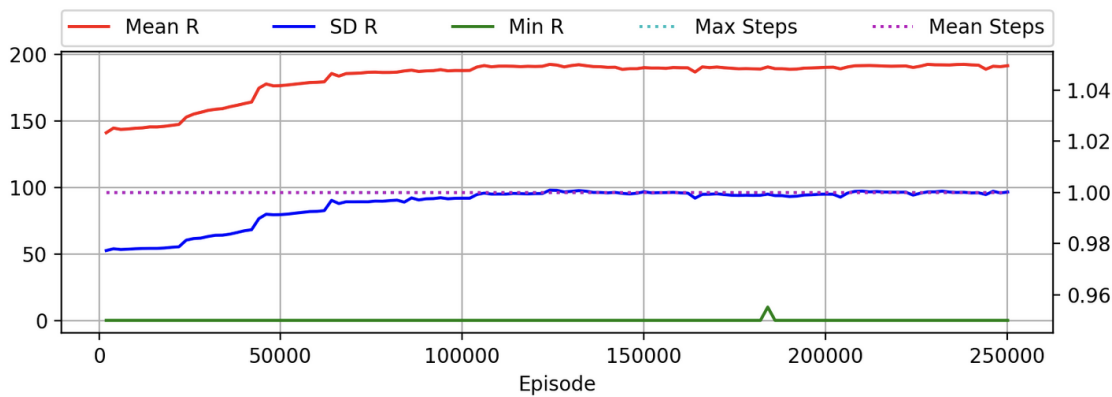

Fig. 1 Learning Curves

### B. Comparison of NLIP, RL and Heuristic

As solving the NLIP was not fast enough for real-time WTA computation, we also developed a simple heuristic to use in our comparison. This heuristic attempts to spread the HSW resources across available targets, but also prioritizes target value. In other words, it is somewhere between assigning all HSW to the most valuable target and an assignment that attempts uniform coverage of all the feasible targets. The heuristic takes into account feasibility and observed target value, but not time to go (which is related to probability of interception).

Table 5 compares the performance of the NLIP benchmark, RL policy, and the heuristic. Performance is quantified by the median value of total value of destroyed targets over all episodes (the "Median Value" column). We use the median rather than mean because the distribution is highly skewed, capped at zero on the low side and with a long tail on the high side. The "% NLIP" column for RL is computed as 100× the ratio of the "Median Value" column for RL and the "Median Value" column for NLIP, and similarly for the heuristic. The comparisons are computed using 30,000 simulated episodes. At the start of each episode, the number of HSW is drawn uniformly from the range (10,20), and the number of targets is drawn from the range(4,12). The cases in column 1 of Table 5 include both the nominal scenario that was used for RL policy optimization as well as additional cases to test generalization performance. These cases are described in Table 6. For each case we also note the empirical mean cumulative probability of an HSW being intercepted. Importantly, for the nominal case, the mean, standard deviation, and maximum computation time (in milliseconds) for the NLIP WTA policy is (2025, 988, 71627), although in other tests the run time has been as high as 265,000 ms[§]. In contrast, the RL WTA policy has mean, standard deviation, and maximum computation times of (1.48, 0.37, 28). The 28 ms outlier may have been due to a rare memory bandwidth bottleneck, as running the RL policy forward should have very predictable computation times. For the nominal case, the mean, standard deviation, minimum, and maximum of the number of feasible targets for an HSW at the time the WTA is computed (i.e., the start of an episode) are 4.1, 1.6, 1.2, and 10.0, respectively.

Table 5   Comparison of NLIP, RL, and heuristic WTA policies

|  | NLIP | RL |  | Heuristic |  |
| --- | --- | --- | --- | --- | --- |
| Case | Median Value | Median Value | % NLIP | Median Value | % NLIP |
| Nominal | 14.18 | 13.05 | 92 | 8.78 | 62 |
| Threat 1 | 17.47 | 15.60 | 89 | 14.41 | 82 |
| Threat 2 | 12.31 | 11.53 | 94 | 6.85 | 56 |
| Sensor Noise | 14.02 | 12.81 | 91 | 8.72 | 62 |
| Threat Targeting | 10.48 | 9.69 | 93 | 6.0 | 57 |
| 25km range | 14.28 | 13.27 | 93 | 8.71 | 61 |

We see that the RL WTA policy gives near-optimal performance with a 1000× improvement in computational efficiency, and generalizes well to novel engagement scenarios. In contrast, although the heuristic policy has computational efficiency close to that of the RL policy, aside from the "Threat Model 1" case, its performance is far from optimal[¶]. The RL policy appears to generalize well, but less so for the reduced threat level in the "Threat Model 1" case and the "Value Distribution" case. We note that the RL WTA performance can likely be increased by experimenting with hyperparameters, including CNN architecture and episodes per rollout. It is also possible that with a higher fidelity engagement model, the RL policy could conceivably outperform the NLIP benchmark. This might occur if there are elements of the higher fidelity model that could only be approximated in the NLIP framework. For example, with a higher fidelity threat model, the probability of interception would depend on the altitude and speed of the HSW, which is path dependent and cannot be represented in closed form as with Eq. 8. Similarly, a higher fidelity sensor model would have accuracy that is a function of distance to target and look angle. A sample engagement with 20 HSW and 12 targets using the RL WTA policy is illustrated in Fig. 2, where the plotted target's diameter is proportional to its value.

The maximum number of HSW and targets in these experiments was chosen as 20 and 12 due to memory limitations of the NLIP solver. Nevertheless, to demonstrate the scalability of RL optimization we conducted experiments where we optimized two policies where these were increased to (40,24) and (60,36). The test results are shown in Table 7, where in each episode, the number of HSW and targets for the (40,24) case is randomly drawn from (20,40) and (8,24), respectively. For the (60,36) case, the number of HSW and targets is randomly drawn from (30,60) and (12,36), respectively. We scaled the size of the expected and actual target regions linearly with the maximum number of threats. Note that the RL run times were independent of the sampled numbers of HSW and targets, whereas the heuristic run times were much longer for the maximum number of HSW and targets, as the heuristic implements two loops, an outer loop over the HSW and an inner loop over the targets. The "Ratio" column gives the ratio of $m \times n$ of the second and third cases to the first (nominal) case. Interestingly, the run times in [8] for the $m = 20, n = 10$ and $m = 40, n = 20$ cases were 11 ms and 25 ms, respectively. In contrast, our run times for the $m = 20, n = 12$ and $m = 40, n = 24$ cases were 2ms and 11 ms, respectively. This is despite the fact that our CPU frequency is 3.2 MHz versus 3.7 MHz for the CPU used in [8]. This could be due to the efficiency of the convolutional front end as opposed to a fully connected network architecture. That said, our network architecture is certainly far from optimal as the maximum engagement size increases, with the largest fully connected layer approaching $10000 \times 20000$ for the (60,36) case. Adding another convolutional layer would mitigate this problem, but the size of the multi-catagorical output layer (which scales linearly with $m \times n$) will be more difficult to reduce.

---

[§]For the "Threat Targeting" case, NLIP hung, and was terminated after 24 hours of computation on the same episode, then restarted to create the results shown in Table 5.

[¶]Note that the heuristic policy occasionally has median values close to whole numbers, we believe this is due to the discretization of target values. We see this for all three policies using the 75 percentile metric

Table 6   Case Descriptions

| Label | Description |
| --- | --- |
| Nominal | This uses the HSW and target initialization parameters given in the problem formulation (Section II), with the number of HSW and targets randomly drawn at the start of each episode as described earlier in this section. Mean cumulative intercept probability of 63% |
| Threat Model 1 | The mean of both the targeting and intercept probabilities are shifted down, reducing the threat level. Specifically, at the start of each episode, the bounds for both the targeting and intercept probabilities are set to (0, 0.25). Mean cumulative intercept probability of 35% |
| Threat Model 2 | The mean of both the targeting and intercept probabilities are shifted up. increasing the threat level. Specifically, at the start of each episode, the bounds for the targeting probability are set to (0.15, 0.25) and those for the intercept probability are set to (0.15, 0.25). Mean cumulative intercept probability of 71% |
| Sensor Noise | We increase the probability of the observed value being shifted up or down one class from 0.1 to 0.2. Mean Cumulative intercept probability of 61% |
| Threat Targeting | Here the probability of an HSW being targeted by a threat is proportional to the value of the target assigned to that HSW. From the defenders perspective, this is a logical strategy, and from the perspective of a WTA policy with no knowledge of this defense strategy, it will result in too few HSW being assigned to high value targets. Mean cumulative intercept probability of 70% |
| 25km Range | The initial range to assigned target is set to 25km. Mean cumulative intercept probability of 61% |

Table 7   Scalability Experiments

| | | | RL | | Heuristic | |
| --- | --- | --- | --- | --- | --- | --- |
| Case | $n \times m$ | Ratio | Median Value | Mean Run Time (ms) | Median Value | Mean Run time (ms) |
| (20,12) | 240 | 1 | 13.1 | 2 | 8.78 | 9 |
| (40,24) | 960 | 4 | 21.8 | 11 | 14.3 | 24 |
| (60,36) | 2160 | 9 | 29.4 | 35 | 20.3 | 45 |

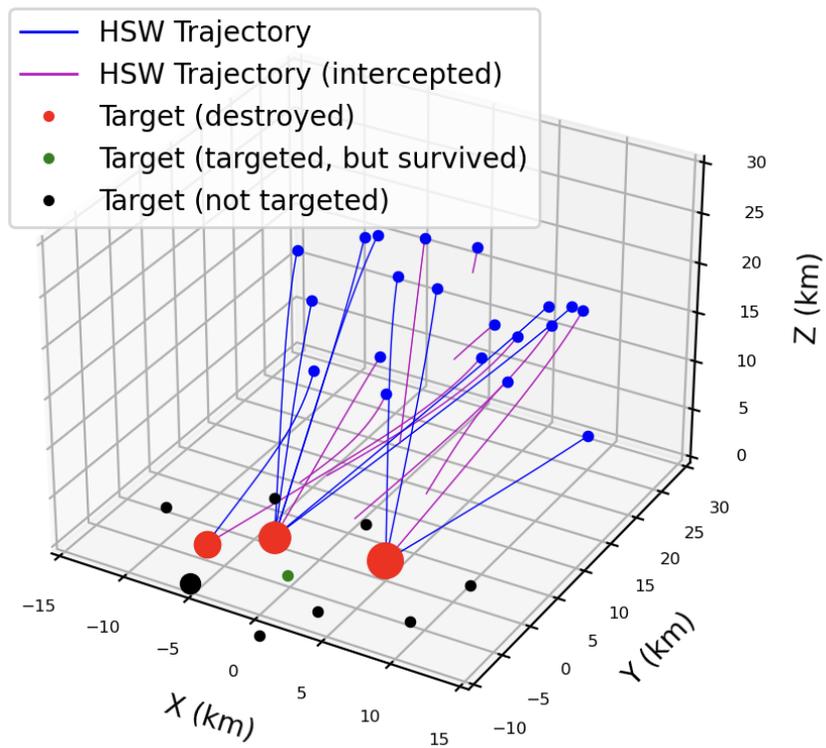

Fig. 2    Sample Engagement

## V. Applying the RL WTA Policy to Dynamic WTA

In Section IV we ran some limited experiments where we applied all three WTA policies to the dynamic WTA problem, but found little difference in performance. We speculated that this was due to our feasibility metric not being applicable later in the engagement. However, it is also possible that in general, dynamically modifying the WTA for a hypersonic strike with closely spaced arrival times will always have limited impact on performance. This is due to the relatively short time of the engagement (8-16 seconds in our simulations), which makes it highly unlikely that re-computing the WTA when a target is destroyed will give sufficient time for a HSW to divert to a new target. Similarly, except for the case where a HSW is intercepted early in the engagement, it is unlikely that another HSW could be diverted to a new target.

Nevertheless, dynamic WTA is also applicable to strikes using multiple waves, i.e., two or more groups of HSW with delayed arrival times between groups, but arrival times within a group closely spaced. In this case, the RL static WTA policy could easily be used to implement dynamic WTA using the same feasibility criterion as in the experiments. Once in sensor range, the second wave would compute its WTA taking into account the current distribution of targets, which would hopefully have changed due to the first wave destroying some fraction of the targets. This could in some cases be more effective than simultaneous arrival of both groups, particularly for the case where the actual threat environment and target distribution differs substantially from the models used in WTA optimization, resulting in inefficient allocation of HSW. Moreover, if the first wave can communicate target and threat information to the second wave, then the second wave could use this information to modify its WTA policy and increase strike effectiveness (and so on for subsequent waves), with the final wave assessing and communicating strike effectiveness of the prior waves to any friendly observer within communication range. Ideally, the second wave would arrive within sensor range once the first wave has completed its strike; this would minimize the time targets have to reload missile launchers.

Finally, we point out that simultaneous arrival of all HSW in a strike group (as opposed to multiple waves of arrivals) is likely overrated in the hypersonic strike application. An adversary's fire control radar likely has a range of around 200 km (similar to our Aegis system), and the targets will have surface to air missiles (SAM) that can engage threats hundreds of kilometers from the launcher. In comparison, due to size and weight constraints, the sensors on an HSW will likely be limited to 30 km or less. And hypersonic vehicles are easily spotted from orbiting satellites due to their heat signature, giving an adversary even more advanced warning. Consequently, there is no element of surprise[6] in such an attack, and the sole advantage of simultaneous arrival rather than multiple waves is limited reload time for the SAM fielded by the adversary. This advantage must be compared to the benefits of staged arrival.

## VI. Conclusion

We demonstrated that RL can be used to optimize an effective WTA policy for a multi-vehicle hypersonic strike against multiple targets. We compared the effectiveness of the RL policy to an optimal NLIP benchmark, and found that RL can provide near optimal performance with a 1000× increase in computational efficiency. The ability to compute a new WTA in real time allows completely autonomous missions to be launched with limited prior information regarding the location and distribution of targets. Moreover, our experiments have shown that the RL policy generalizes well to scenarios outside the bounds of the optimization distribution. Importantly, the RL WTA gives near linear scaling of computation time with the number of hypersonic strike weapons and targets. We also outlined an inter-agent communications framework that allows multiple strike vehicles to maintain a consistent view of the engagement state. Finally, we discussed how the RL WTA policy could be applied to a dynamic WTA problem with multiple waves of attackers. Future work will increase the fidelity of the sensor and threat models, refine the inter-agent communication protocol, and develop an improved target feasibility metric to use with proportional navigation.